\documentclass[runningheads]{llncs}

\usepackage{eccv}

\usepackage{eccvabbrv}

\usepackage{graphicx}
\usepackage{booktabs}
\usepackage{multirow}   % \multirow
\usepackage{array}
\usepackage{makecell}
\usepackage{bm}
\usepackage{wrapfig}
\usepackage{colortbl}
\usepackage{pifont}

\newcommand{\cmark}{\ding{51}} % 对勾
\newcommand{\xmark}{\textcolor{gray!40}{\ding{55}}} 

\definecolor{my_blue}{HTML}{d3eaf2}
\definecolor{my_green}{RGB}{209, 229, 230}
\definecolor{Green}{rgb}{0.85882353, 0.90980392, 0.84705882}
\definecolor{rose}{rgb}{0.60392157, 0.53333333, 0.43921569}
\definecolor{dred}{rgb}{0.7254902, 0.09803922, 0.10588235}

\def\eg{\emph{e.g}\onedot}

\usepackage[accsupp]{axessibility}  % Improves PDF readability for those with disabilities.

\usepackage{hyperref}

\usepackage{orcidlink}

\begin{document}

\title{FlowCIR: Semantic Transport via Flow Matching for Zero-Shot Composed Image Retrieval} 

\titlerunning{FlowCIR}

\author{
Zhenqi He\textsuperscript{1}\orcidlink{0009-0000-2265-7159} \and 
Ziqi Jiang\textsuperscript{1}\orcidlink{} \and 
Yuanpei Liu\textsuperscript{2}\orcidlink{0009-0008-6144-6547} \and 
Yanghao Wang\textsuperscript{1}\orcidlink{} \and 
Teng Wang\textsuperscript{2}\orcidlink{} \and 
Long Chen\textsuperscript{1}\textsuperscript{\dag}\orcidlink{0000-0001-6148-9709}
}

\authorrunning{He et al.}

\institute{The Hong Kong University of Science and Technology, Hong Kong SAR
\\
\and
The University of Hong Kong, Hong Kong SAR\\
\email{zheci@connect.ust.hk, longchen@ust.hk}}

\maketitle
\renewcommand{\thefootnote}{}
\footnotetext{\textsuperscript{\dag} Corresponding author.}

\begin{abstract}
Zero-shot composed image retrieval (ZS-CIR) aims to retrieve a target image by editing a reference image with a natural-language instruction, without relying on domain-specific annotated triplets.
Most existing ZS-CIR methods rely on textual inversion to translate the reference image into pseudo-text tokens and then compose them with the instruction via simple concatenation in the text space, which can be lossy and brittle for fine-grained semantics.
In this work, we propose a new paradigm, namely \textsc{\textbf{FlowCIR}}, that casts ZS-CIR as \emph{conditional semantic transport} between reference and target embeddings.
Leveraging \emph{conditional flow matching}, our model learns a lightweight transport field that maps the instruction representation toward a target-aligned query embedding conditioned on the reference image.
Since FlowCIR operates on pre-extracted VLM embeddings and trains only a small transport module without updating the image or text encoder, it offers a computationally efficient training protocol compared with prior textual-inversion-based approaches.
% The resulting framework is training-efficient, requiring roughly $10\times$ fewer training resources than prior textual-inversion-based approaches.
We further identify negation and removal as a major failure mode of VLM-based composition.
To address this, we propose an inference-only Multi-Negative Steering strategy that steers a negation-containing relative instruction away from its negated semantics, mitigating the limited negation handling of VLMs and improving robustness on negation-heavy queries.
Extensive experiments on standard CIR benchmarks demonstrate that FlowCIR achieves strong and competitive performance compared with recent ZS-CIR methods.
Project page: \url{https://hkust-longgroup.github.io/FlowCIR}
  
\end{abstract}
\section{Introduction}
\begin{figure}[t]
  \centering
  \includegraphics[width=1\linewidth]{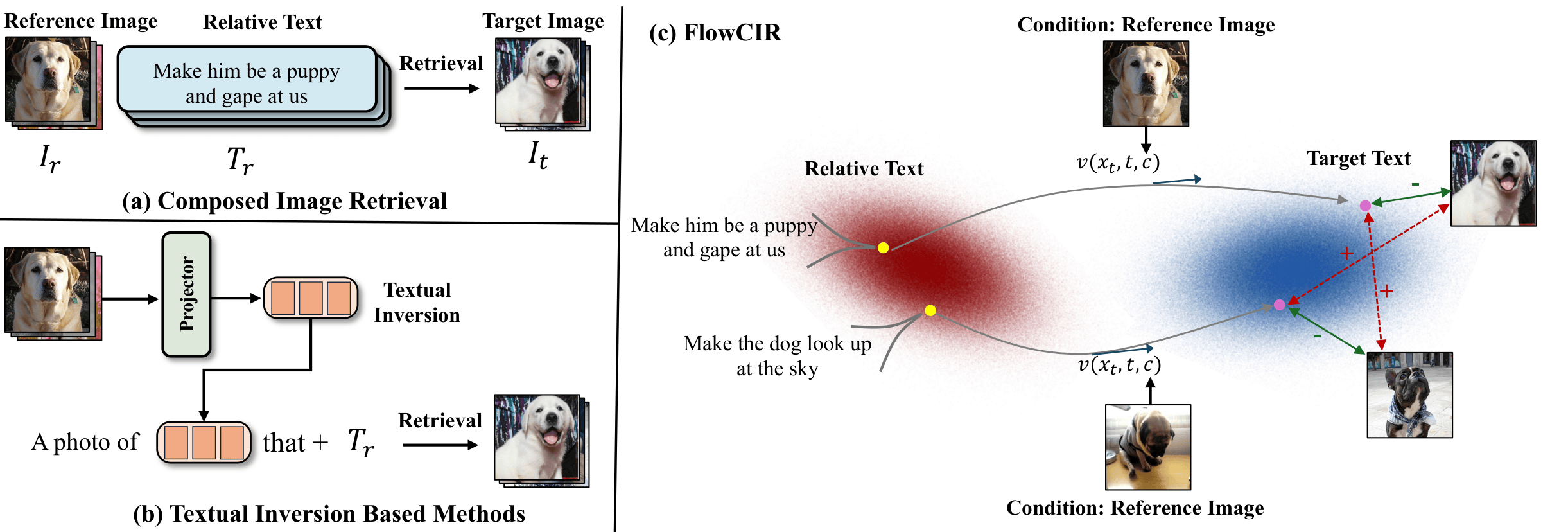}
  \vspace{-8pt}
  \caption{
(a) Illustration of Composed Image Retrieval.
(b) Prior methods are composed by textual inversion and token-level fusion in text space. 
(c) \textsc{FlowCIR} composes via conditional flow matching to produce a target-oriented retrieval query.}
% \vspace{-8pt}
  \label{fig:teaser}
\end{figure}

Composed image retrieval (CIR)~\cite{zhan2025elip,duan2025fuzzy,Combiner,ARTEMIS,CIRR,FashionVLP,vo2019composing} offers a practical visual search interface: rather than describing a target image from scratch, a user provides a reference image and a relative instruction, and the system retrieves images that match the intended edit (see Fig.~\ref{fig:teaser}(a)).
Previous approaches~\cite{Combiner,tang2026heterogeneousuncertaintyguidedcomposedimage} are supervised and rely on human-annotated triplets 
%(inference image $I_r$, relative text $T_r$ and target image $I_t$)
whose large-scale collection is laborious and time-consuming.
In addition, models trained on a particular dataset or domain often generalize poorly beyond their training distribution.
These limitations motivate the zero-shot setting~\cite{baldrati2023zero,saito2023pic2word}, which studies CIR without domain-specific annotations and typically relies on unsupervised objectives or synthetic supervision to learn more transferable representations.

\setlength{\intextsep}{0.2pt}      % 图与上下文字的竖直间距
 \setlength{\columnsep}{6pt}      % 图与文字的水平间距（关键）
 \begin{wrapfigure}[15]{r}{0.45\textwidth}
   % \vspace{-0.6\baselineskip}
   \centering
   \includegraphics[width=0.45\textwidth]{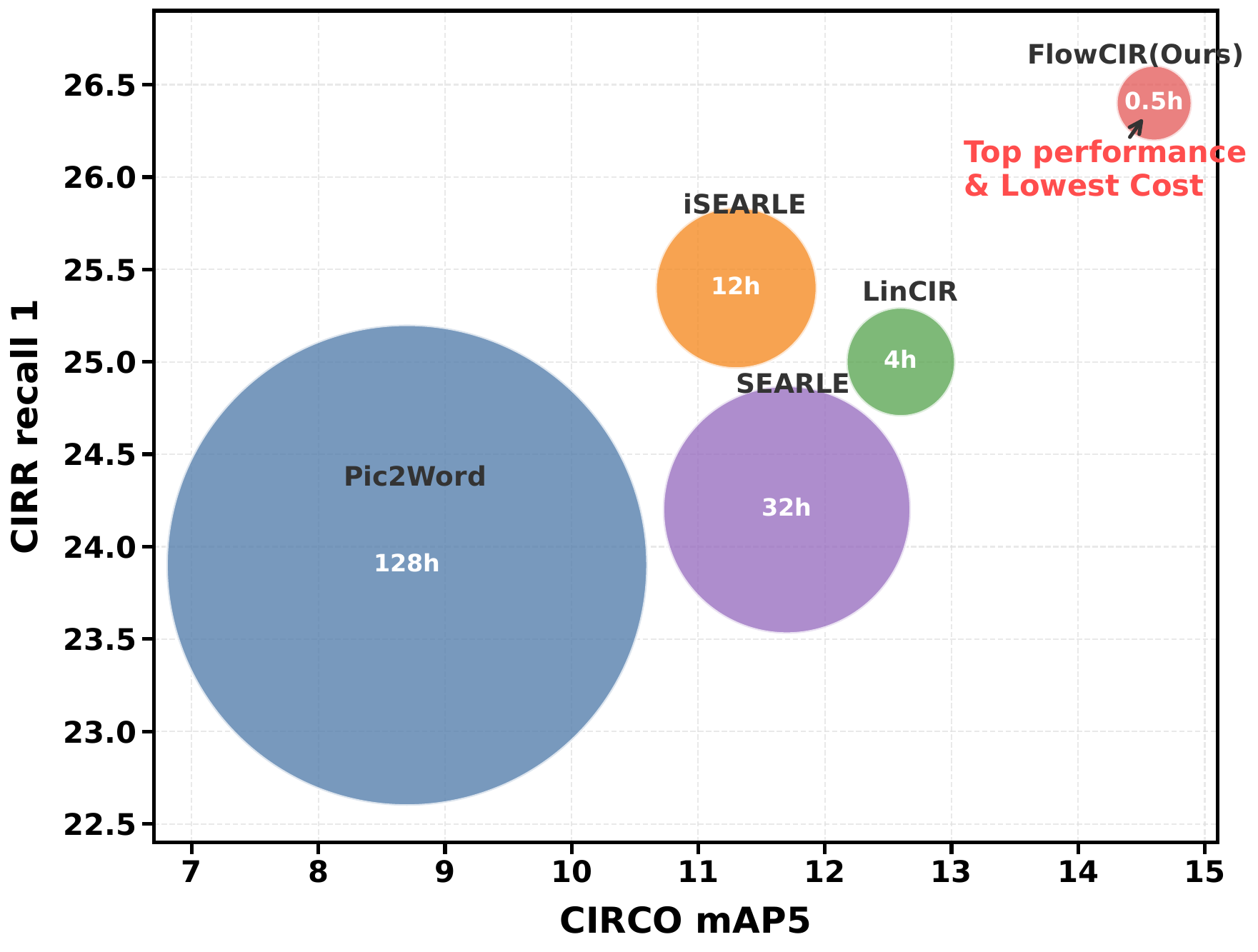}
   % \vspace{-12pt}
   \caption{Performance on CIRCO (mAP5) and CIRR (Recall1) with training hours (in single GPU) as bubble size.}
   \label{fig:bubble}
 \end{wrapfigure}
 
Recent zero-shot CIR methods~\cite{saito2023pic2word, baldrati2023zero,gu2024language, tang2024context} largely build on textual inversion over well-aligned vision--language models~\cite{clip,blip}, effectively reducing cross-modal composition to a text-only manipulation problem.
Concretely, they learn a projector to \emph{transfer} the reference image $I_r$ into a small set of pseudo-text tokens and compose them with the relative instruction $T_r$ in the text space to form a retrieval query, which is then matched against gallery image embeddings as shown in Fig.~\ref{fig:teaser}(b).
This paradigm rests on the key assumption that the semantics of an image can be faithfully captured by a few textual tokens and manipulated entirely within a textual space.
However, this assumption is often impractical: compressing a rich reference image into only a handful of tokens inevitably loses fine-grained visual cues, so the composed representation may miss essential semantics needed for accurate retrieval.
Moreover, learning an effective inversion projector often requires substantial computation, typically involving hours of large-batch contrastive training on multiple GPUs with web-scale image--text data~\cite{saito2023pic2word,gu2024language,gu2023compodiff}.
These limitations motivate a more reliable and training-efficient mechanism for cross-modal composition beyond textual inversion.

In this paper, we reframe zero-shot CIR as a \emph{semantic transport} problem between the reference and target distributions in the embedding space of vision--language models, and accordingly employ flow matching to bridge the two distributions.
Building on this view, we introduce \textbf{\textsc{FlowCIR}}.
Formally, \textsc{FlowCIR} learns a conditional transport that maps the relative-instruction text embedding to a target-oriented text embedding, conditioned on the reference image embedding.
Different from prior methods that form retrieval queries via static token operations in text space, we perform composition as a \emph{continuous} semantic transport through \emph{conditional flow matching}, which better preserves semantics and enables a progressive, reference-guided composition process.
This formulation provides a principled mechanism for cross-modal composition and is also training-efficient: we only learn a lightweight conditional velocity field, enabling training within about one hour on a single GPU.
As illustrated in Fig.~\ref{fig:bubble}, our method achieves strong performance on both CIRR and CIRCO while incurring substantially lower training cost than prior baselines.

Meanwhile, as our transport is initialized from the instruction embedding, it may be influenced by a known weakness of CLIP-style VLMs: negation is often geometrically ambiguous in the text space~\cite{alhamoud2025vision,ranjbar2025spacevlm}.
For example, expressions such as ``a dog'' and ``remove/no dog'' may lie undesirably close, providing an unreliable starting signal and biasing the transfer toward the negated concept.
%
% Therefore, inspired by \cite{ranjbar2025spacevlm}, we propose a \emph{multi-negative aware} strategy: we identify several undesired semantic anchors implied by the instruction and add an explicit repulsion term during transport, which helps separate affirmative and negated semantics and improves retrieval on negation-heavy queries.
Therefore, inspired by~\cite{ranjbar2025spacevlm}, we propose an inference-only \emph{multi-negative steering} strategy that explicitly pushes the composed instruction representation away from negative semantic directions, alleviating affirmative/negated collapse in CLIP space and improving retrieval on negation-heavy queries.

% \lc{the relation with the previous paragraph is not very smooth} However, even with a more principled transport-based composition, a key challenge remains in zero-shot CIR: negation often induces ambiguous geometry in VLM space~\cite{alhamoud2025vision,ranjbar2025spacevlm}.
% For example, the text embeddings of ``dog'' and ``no dog'' can be undesirably close in the text space of CLIP, making the initial direction unreliable and causing the transported representation to drift toward the negated concept rather than away from it.
% %
% Motivated by evidence that explicitly modeling negatives can better steer vision--language representations~\cite{ranjbar2025spacevlm}, we introduce a multi-negative aware strategy that disentangles negated semantics from their affirmative counterparts.
% %
% Concretely, given a negation-containing instruction, we identify multiple undesired semantic anchors and repel the transported embedding from their associated regions in the embedding space, preventing positive/negative collapse and effectively improving retrieval accuracy on negation-heavy queries.

In summary, our contributions are threefold. (i) We introduce \textsc{\textbf{FlowCIR}}, a new paradigm for zero-shot composed image retrieval by conditional semantic transport in embedding space. (ii) We propose a multi-negative steering strategy that substantially improves robustness to negation and removal instructions, a prominent failure mode of CLIP-based composition.
(iii) We achieve consistent improvements on standard CIR benchmarks, while remaining training-efficient.
\section{Related Work}
\noindent
\textbf{Zero-Shot Composed Image Retrieval.}
% Composed image retrieval (CIR)~\cite{vo2019composing, yu2020curlingnet} aims to retrieve target images given a reference image and a modification text describing the desired change. To avoid relying on heavily annotated real-world triplets, recent studies propose zero-shot CIR (ZS-CIR)~\cite{saito2023pic2word, baldrati2023zero}, where models perform composition and retrieval without task-specific manual annotations. A critical challenge in ZS-CIR is how to effectively compose the visual evidence from the reference image with the semantic shift implied by the text.
Existing ZS-CIR methods can be broadly grouped into three lines. (1) {\textit{Textual-inversion based methods}}~\cite{saito2023pic2word, baldrati2023zero,gu2024language, tang2024context,byun2025efficientposthocframeworkreducing} convert image information into learnable pseudo tokens, and then perform composition in the text embedding space for retrieval; representative works include Pic2Word~\cite{saito2023pic2word} and SEARLE~\cite{baldrati2023zero} and subsequent variants that design context-dependent mappings~\cite{tang2024context} or multiple pseudo words~\cite{du2024image2sentence} to enhance controllability. (2) \textit{Generative-based methods} introduce generative models (\eg, diffusion models) to synthesize target-oriented features~\cite{gu2023compodiff} or leverage large-scale pretrained image generators (\eg, Stable Diffusion~\cite{rombach2022high}) to produce pseudo target images as auxiliary visual evidence~\cite{wang2025generative}.  (3) \textit{MLLM-based training-free methods}~\cite{dualCIR,Sun_2025_ICCV,karthik2023vision} leverage large vision--language models to obtain richer textual signals without task-specific training: they first caption the reference image with a VLM and then use an LLM/MLLM to parse, rewrite, or decompose the modification instruction and fuse it with the caption to form a composed query for retrieval~\cite{yang2024ldre, li2024improving, karthik2023vision}, at the cost of extra inference-time computation.

In contrast to the above paradigms, we formulate composed image retrieval as \emph{conditional semantic transport} and instantiate composition with conditional flow matching, which directly fuses the reference image evidence and relative instruction via a learned velocity field rather than token inversion, image synthesis, or inference-time prompting. Moreover, unlike DualCIR~\cite{dualCIR} and CoTMR~\cite{Sun_2025_ICCV} that leverage positive/negative semantics mainly by constructing separate textual queries and fusing image-text similarity scores, the Multi-Negative Steering aims to reshape the instruction embedding by steering it away from negative semantics.

% We propose to use conditional flow matching as an information-composition mechanism for composed image retrieval, modeling the fusion of reference image evidence and modification text as a conditional transport process in the embedding space.

\noindent 
\textbf{Flow Matching (FM).} FM learns a time-conditioned velocity field (a neural ODE) that transports samples along a predefined probability path from a simple prior to the data distribution by directly regressing the corresponding velocities~\cite{lipman2023flow,albergo2023building,liu2022flow}.
Early studies~\cite{esser2024scaling,geng2025mean} instantiate this transport from noise to images, achieving high-fidelity synthesis on par with diffusion models~\cite{ho2020denoising} while retaining simpler training and sampling dynamics.
More recent text-to-image FM variants~\cite{liu2025flowing,he2025flowtok} further learn cross-modal transports from text embeddings to image embeddings, enabling direct language-to-vision generation through learned velocity fields.
Beyond text-conditioned synthesis, FM has also been adapted to visual domain transfer problems such as monocular depth estimation~\cite{gui2025depthfm} and semantic segmentation~\cite{bogensperger2025flowsdf,wang2025deforming}, by formulating them as transport between distributions defined over different visual representations.

\noindent 
\textbf{Generative Models for Perception.}
Beyond synthesis, modern generative models have become increasingly useful for visual recognition and perception tasks~\cite{wang2025inversion,wang2025noise,liu2025generate,clark2023text,li2023your,chen2023robust,qi2024simple,wang2026lisalikelihoodscorealignment}. One prominent direction leverages generators as \emph{data engines}, producing realistic and diverse samples to mitigate limited supervision and improve performance in fine-grained recognition~\cite{islam2024diffusemix,fu2024dreamda}, few-shot learning~\cite{wang2025inversion}, long-tailed classification~\cite{koh2025synthetic}, and category discovery~\cite{liu2025generate}.
In parallel, a complementary line uses the generative process for recognition itself: text-to-image diffusion models can be repurposed as zero-shot or few-shot classifiers by evaluating class-conditioned denoising likelihoods~\cite{clark2023text,li2023your,chen2023robust}, with efficiency improved via hierarchical prompting~\cite{ning2024hierarchical} and auxiliary image encoders~\cite{qi2024simple}. More recent studies further explore parameterizing few-shot adaptation through diffusion time steps to capture subtle attributes~\cite{yue2024few}, and selecting or optimizing the injected noise to stabilize and strengthen the recognition pipeline~\cite{wang2025noise}.
FMA~\cite{jiang2025exploring} and HFM~\cite{li2026pathdecoupledhyperbolicflowmatching} employ unconditional flow matching to bridge image and text within VLM embedding space for few-shot learning, and FlowComposer~\cite{He2026FlowComposer} employs flow matching for compositional zero-shot learning.

In contrast to the above methods, we are the first to explore conditional flow matching for composed image retrieval.
\section{Method}
\subsection{Preliminaries}
\textbf{Flow Matching.} Flow Matching~\cite{lipman2023flow,albergo2023building} aims to learn a continuous-time transport between two distributions ($X_0 \rightarrow X_1$) by regressing a time-dependent velocity field.
Under the rectified flow formulation~\cite{liu2022flow}, a linear path is defined between paired samples $(\bm{x}_0\sim X_0,\bm{x}_1\sim X_1)$ as $\bm{x}_t=(1-t)\bm{x}_0+t\bm{x}_1, t\in[0,1]$, 
whose ground-truth velocity is constant: $\bm{v}^{\star}(\bm{x}_t,t)=\bm{x}_1-\bm{x}_0$.
A neural velocity field $\bm{v}_{\theta}(\bm{x}_t,t)$ is then trained by minimizing
\begin{equation}
\mathcal{L}_{\mathrm{FM}}(\theta)
=
\mathbb{E}_{\bm{x}_0,\bm{x}_1,t}
\Bigl[
\bigl\|
\bm{v}_{\theta}(\bm{x}_t,t)
-
(\bm{x}_1-\bm{x}_0)
\bigr\|_2^2
\Bigr],
\end{equation}
which is equivalent to matching the marginal velocity field in the original flow-matching formulation~\cite{lipman2023flow,liu2022flow}.
Once trained, transport is performed by solving the ODE $\frac{d\bm{x}_t}{dt}=\bm{v}_\theta(\bm{x}_t,t)$
from $t=0$ to $t=1$.
Conditional flow matching extends this formulation by conditioning the velocity field on the condition $\bm{c}$, yielding
\begin{equation}
\mathcal{L}^{(\mathrm{FM})}_{\mathrm{cond}}(\theta)
=
\mathbb{E}
\Bigl[
\bigl\|
\bm{v}_\theta(\bm{x}_t,t,\bm{c})
-
(\bm{x}_1-\bm{x}_0)
\bigr\|_2^2
\Bigr].
\end{equation}
In our setting, we adopt this conditional formulation to transport the relative instruction embedding toward the target text embedding, conditioned on the reference image embedding.

\noindent \textbf{Problem Setup and Notation.}
Given a gallery of candidate images $\mathcal{D}=\{I_i\}_{i=1}^{N}$, zero-shot composed image retrieval (ZS-CIR) considers a query triplet $\langle I_r, T_r, I_t\rangle$, where $I_r$ is a reference image, $T_r$ is a relative text instruction describing the desired modification to $I_r$, and $I_t$ is the target image satisfying the instruction.
The goal is to compose the visual content of $I_r$ with the semantics of $T_r$ to retrieve $I_t$ from $\mathcal{D}$.
In the zero-shot setting, training does not use CIR-specific triplets from the target benchmarks, but instead relies on external supervision such as large-scale image--text data or synthesized tuples.

In the following, given $\langle I_r, T_r, I_t\rangle$, we denote a target-side text description associated with $I_t$ as $T_t$.
We encode both images and texts into a shared VLM embedding space, and denote all embeddings by $\bm{x}$.
Using a pretrained image encoder $E_I(\cdot)$ and text encoder $E_T(\cdot)$, we obtain $\ell_2$-normalized features
$\bm{x}_{I_r}=E_I(I_r)/\|E_I(I_r)\|$,
$\bm{x}_{T_r}=E_T(T_r)/\|E_T(T_r)\|$,
$\bm{x}_{T_t}=E_T(T_t)/\|E_T(T_t)\|$,
and $\bm{x}_{I_t}=E_I(I_t)/\|E_I(I_t)\|$.
Here, $\bm{x}_{I_r}$, $\bm{x}_{T_r}$, $\bm{x}_{T_t}$, and $\bm{x}_{I_t}$ denote the embeddings of the reference image, relative instruction, target-side text, and target image.

\subsection{Conditional Flow Matching for Semantic Transport}
\label{subsec:cfm}
We train in a zero-shot manner using synthesized compositions of the form $\langle I_r, T_r, T_t, I_t\rangle$, where $I_r$ is a reference image, $T_r$ is a relative instruction, and $(T_t, I_t)$ describes the target side.
% Given such a tuple, our goal is to learn a composed query embedding that retrieves $I_t$ from the gallery.
% We extract frozen, $\ell_2$-normalized vision--language embeddings with a pretrained image encoder $E_I(\cdot)$ and a text encoder $E_T(\cdot)$ to obtain
% $\bm{x}_{I_r}=\dfrac{E_I(I_r)}{\|E_I(I_r)\|}\in\mathbb{R}^d$,
% $\bm{x}_{T_r}=\dfrac{E_T(T_r)}{\|E_T(T_r)\|}\in\mathbb{R}^d$,
% $\bm{x}_{T_t}=\dfrac{E_T(T_t)}{\|E_T(T_t)\|}\in\mathbb{R}^d$,
% and $\bm{x}_{I_t}=\dfrac{E_I(I_t)}{\|E_I(I_t)\|}\in\mathbb{R}^d$, where $d$ and $\|\cdot\|$ refer to the feature dimension and the $l2$ normalization.
% where $\bm{x}_{I_r}$ serves as the conditioning signal, $\bm{x}_{T_r}$ is the source instruction embedding, $\bm{x}_{T_t}$ is the target-text destination, and $\bm{x}_{I_t}$ provides retrieval supervision.
%
Our model then learns a conditional semantic transport mapping $(\bm{x}_{T_r}, \bm{x}_{I_r}) \mapsto \bm{x}_{T_t}$ within the embedding space.
We utilize conditional flow matching to model composition as semantic transport.
Given a tuple and the extracted features, we sample a time scalar $t\sim\mathcal{U}(0,1)$ and construct the interpolated state along a linear path as $\bm{x}_t = (1-t)\bm{x}_{T_r} + t\,\bm{x}_{T_t}$.
% \begin{equation}
% \bm{x}_t = (1-t)\bm{x}_{T_r} + t\,\bm{x}_{T_t}.
% \end{equation}
Our flow matching network $f_\theta$ predicts a time-dependent velocity conditioned on the reference image $\hat{\bm{v}}_\theta = f_\theta(\bm{x}_t,t,\bm{x}_{I_r})$, 
% \begin{equation}
% \hat{\bm{v}}_\theta = f_\theta(\bm{x}_t,\bm{x}_{I_r},t),
% \end{equation}
and is trained to match the ground-truth constant velocity $\bm{v}^\star=\bm{x}_{T_t}-\bm{x}_{T_r}$ via the flow-matching regression objective: 
\begin{equation}
\mathcal{L}_{\mathrm{FM}}
= \mathbb{E}\!\left[\left\|\hat{\bm{v}}_\theta(\bm{x}_t,t,\bm{x}_{I_r})-(\bm{x}_{T_t}-\bm{x}_{T_r})\right\|_2^2\right].
\end{equation}
Besides regressing the velocity, we adopt a contrastive retrieval objective where the predicted text embedding and the ground-truth image embedding $(\hat{\bm{x}}_{T_t}, \bm{x}_{I_t})$ form a positive pair, and we mine top-$K$ hard negatives from the remaining target-image embeddings in the minibatch to sharpen discrimination against challenging distractors.
We first form a predicted target text embedding $\hat{\bm{x}}_{T_t}$ by a one-step transport from time $t$ to $1$: $\hat{\bm{x}}_{T_t}=\bm{x}_t + (1-t)\,\hat{\bm{v}}_\theta(\bm{x}_t,t,\bm{x}_{I_r})$, 
% \begin{equation}
% \hat{\bm{x}}_{T_t}=\bm{x}_t + (1-t)\,\hat{\bm{v}}_\theta(\bm{x}_t,t,\bm{x}_{I_r}),
% \end{equation}
and compute a sampled InfoNCE loss over the positive and the selected hard negatives.
Specifically, we define the hard-negative index set as the top-$K$ most similar candidates to the query (excluding the positive):
\begin{equation}
\mathcal{H}_K(\hat{\bm{x}}_{T_t})
=\operatorname{TopK}\!\left(
\left\{\,
\Big\langle \tfrac{\hat{\bm{x}}_{T_t}}{\|\hat{\bm{x}}_{T_t}\|},\, \bm{x}^{\,j}_{I_t}\Big\rangle
\,\right\}_{j\neq i}
\right),
\end{equation}
where $\operatorname{TopK}(\cdot)$ returns the indices of the $K$ largest similarities in the minibatch.
The resulting top-$K$ InfoNCE objective is
\begin{equation}
\mathcal{L}_{\mathrm{RET}}
= -\log
\frac{\exp\!\left(\big\langle \tfrac{\hat{\bm{x}}_{T_t}}{\|\hat{\bm{x}}_{T_t}\|},\, \bm{x}_{I_t}\big\rangle/\tau\right)}
{\exp\!\left(\big\langle \tfrac{\hat{\bm{x}}_{T_t}}{\|\hat{\bm{x}}_{T_t}\|},\, \bm{x}_{I_t}\big\rangle/\tau\right)
+\sum_{j\in\mathcal{H}_K(\hat{\bm{x}}_{T_t})}\exp\!\left(\big\langle \tfrac{\hat{\bm{x}}_{T_t}}{\|\hat{\bm{x}}_{T_t}\|},\, \bm{x}^{\,j}_{I_t}\big\rangle/\tau\right)},
\end{equation}
where $\bm{x}_{I_t}^{\,j}$ denotes the target-image embedding of the $j$-th sample in the minibatch, $\tau$ is a temperature scalar, and $\mathcal{H}_K(\hat{\bm{x}}_{T_t})$ indexes the $K$ hardest negatives.
The overall training objective is $\mathcal{L}=\mathcal{L}_{\mathrm{FM}}+\lambda\,\mathcal{L}_{\mathrm{RET}}$, 
% \begin{equation}
% \mathcal{L}=\mathcal{L}_{\mathrm{FM}}+\lambda\,\mathcal{L}_{\mathrm{RET}},
% \end{equation}
where $\lambda$ balances the learning of FM and retrieval.
During inference, we adopt the same one-step transport scheme for efficiency: predicting the velocity $f_\theta(\bm{x}_{T_r},\bm{x}_{I_r},0)$ yields direct one-step transport to $\hat{\bm{x}}_{T_t}$, which is then used to retrieve images by nearest-neighbor search in the gallery embedding space.

\begin{figure}[t]
  \centering
  \includegraphics[width=1\linewidth]{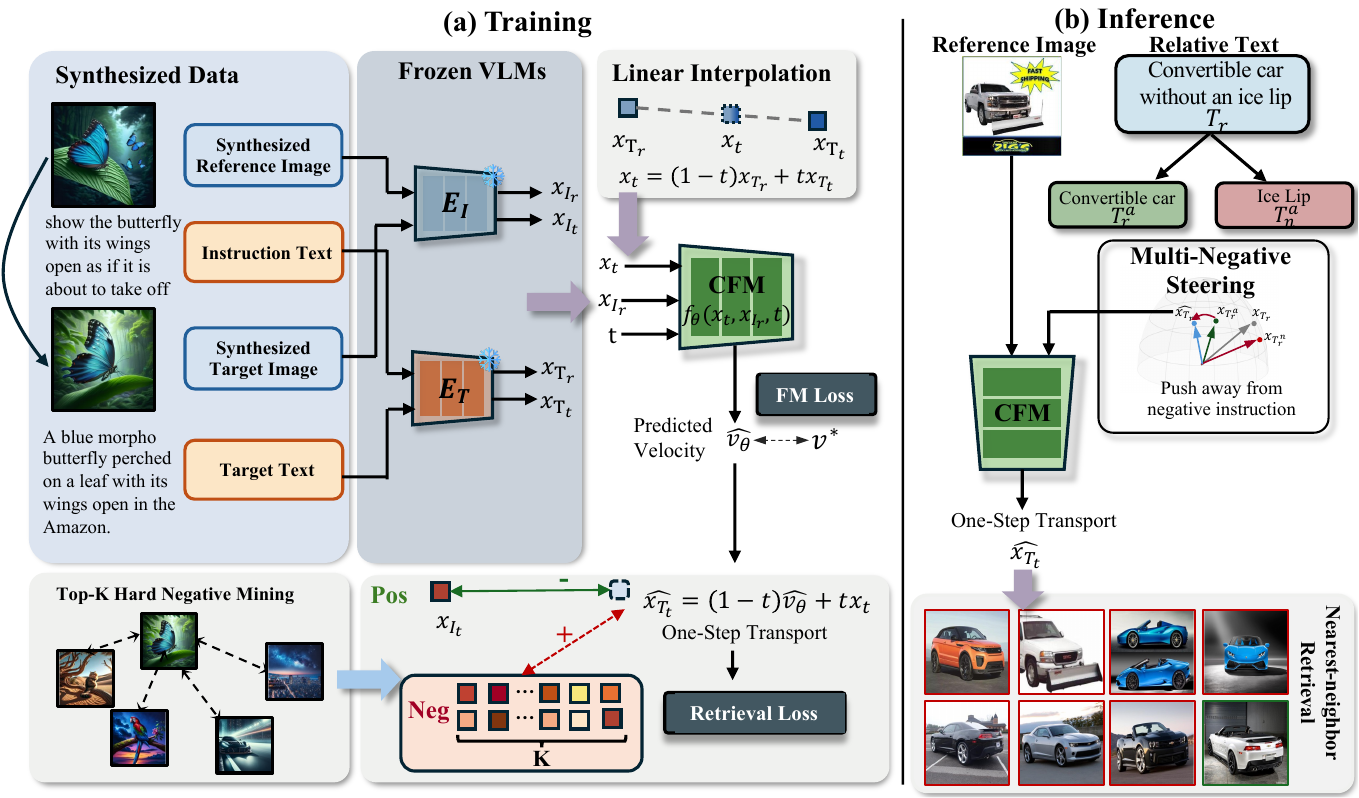}
  \caption{
Framework overview of \textsc{FlowCIR}.
(a) In training, \textsc{FlowCIR} learns a conditional flow-matching transport from relative-instruction embeddings to target-oriented text embeddings under the reference-image condition, together with top-$K$ hard-negative retrieval supervision.
(b) In inference, an inference-only Multi-Negative Steering module adjusts negation-containing instructions before one-step transport for target image retrieval.}
  \label{fig:method}
\end{figure}

\subsection{Multi-Negative Steering Strategy}
\label{subsec:multineg}
Negation-heavy instructions (e.g., \emph{``no dog''}, \emph{``remove stripes''}) often induce ambiguous geometry in contrastive text spaces, where affirmative and negated descriptions can collapse to nearby directions~\cite{alhamoud2025vision}.
To address this issue without additional training, we propose an inference-only multi-negative steering strategy.
During inference, given a relative instruction $T_r$, we decompose it into an affirmative intent $T_r^{a}$ and a set of negated concepts $\{T_{r,k}^{n}\}_{k=1}^{K}$ using a lightweight parser (\eg, a small LLM with simple rule-based heuristics), and encode them as normalized embeddings $\bm{x}_{T_r^{a}}=E_T(T_r^{a})/\|E_T(T_r^{a})\|$ and $\bm{x}_{T_{r,k}^{n}}=E_T(T_{r,k}^{n})/\|E_T(T_{r,k}^{n})\|$.
Our goal is to construct a \emph{steered instruction} embedding $\hat{\bm{x}}_{T_r}$ that remains aligned with $\bm{x}_{T_r^{a}}$ while being repelled from all $\{\bm{x}_{T_{r,k}^{n}}\}_{k=1}^{K}$.

Following SpaceVLM~\cite{ranjbar2025spacevlm}, we model each concept as a hyper-spherical cap in the embedding space: for a normalized point $\bm{x}$ and a cosine threshold $\delta\in[-1,1]$, the cap region is defined as $\mathcal{R}(\bm{x})=\{\bm{z}\in\mathbb{R}^d \mid \bm{x}^\top \bm{z}\ge \delta\}.$
% \begin{equation}
% \mathcal{N}(\bm{x})=\{\bm{z}\in\mathbb{R}^d \mid \bm{x}^\top \bm{z}\ge t\}.
% \end{equation}
Accordingly, a feasible negation-aware representation should lie close to the affirmative intent but outside the neighborhoods induced by negatives, yielding
\begin{equation}
\mathcal{R}(T_r)=\mathcal{R}(\bm{x}_{T_r^{a}})\cap\bigcap_{k=1}^{K}\mathcal{R}^c(\bm{x}_{T_{r,k}^{n}}),
\end{equation}
where $\mathcal{R}^c(\cdot)$ denotes the complement.
To obtain a single embedding compatible with our transport model, we extend this construction by computing, for each negative $\bm{x}_{T_{r,k}^{n}}$, a closed-form ``center'' direction:
\begin{equation}
\tilde{\bm{d}}_k
=
\bm{x}_{T_r^{a}}\frac{\sin\!\left(\alpha+\tfrac{\theta_k}{2}\right)}{\sin\theta_k}
-
\bm{x}_{T_{r,k}^{n}}\frac{\sin\!\left(\alpha-\tfrac{\theta_k}{2}\right)}{\sin\theta_k},
\qquad
\bm{d}_k=\frac{\tilde{\bm{d}}_k}{\|\tilde{\bm{d}}_k\|},
\end{equation}
where $\alpha=\arccos(\delta)$ is set by a cosine-margin hyperparameter $\delta\in[-1,1]$ controlling the hyper-spherical cap region size, and $\theta_k=\arccos(\bm{x}_{T_r^{a}}^\top \bm{x}_{T_{r,k}^{n}})$.
When $\theta_k$ is very small, we use $\tilde{\bm{d}}_k=\bm{x}_{T_r^{a}}-\bm{x}_{T_{r,k}^{n}}$ for numerical stability.
We then aggregate and normalize these directions to form the steered instruction embedding:
\begin{equation}
\hat{\bm{x}}_{T_r}
=
\frac{1}{K}\sum_{k=1}^{K}\bm{d}_k,
\qquad
\hat{\bm{x}}_{T_r}\leftarrow \frac{\hat{\bm{x}}_{T_r}}{\|\hat{\bm{x}}_{T_r}\|},
\end{equation}
and replace the original instruction embedding with $\hat{\bm{x}}_{T_r}$ in our transport pipeline, such that the composed representation remains aligned with the affirmative semantics while being repelled from all negative directions in the embedding space.

\subsection{Theoretical Justification}
\label{subsec:theory}
We provide a theoretical justification for applying conditional flow matching for zero-shot CIR. Formally, standard flow matching is designed for distribution-level transport, where reaching any valid target mode is sufficient, whereas composed image retrieval requires \emph{instance-specific semantic alignment}: for a given pair $(x_{I_r},x_{T_r})$, the model should recover its corresponding target $x_{T_t}$ rather than an arbitrary mode. Below, we analyze (1) why conditioning on the reference image is necessary, and (2) when the learned conditional velocity yields exact point-to-point semantic transport.

\noindent \textbf{Ambiguity of Unconditional Flow.} 
% In a real-world dataset, a single relative instruction (\eg, ``change to red'') is highly ambiguous. Without observing the visual context, an instruction feature $x_{T_r}$ maps to a mixture of multiple targets $\{x_{T_t}^k\}_{k=1}^K$, where $K$ is the number of potential targets.
In realistic composed retrieval data, a single relative instruction $x_{T_r}$ (\eg, Make the dog look up at the sky) can correspond to multiple valid targets $\{I_t^k\}_{k=1}^K$.
As a result, without conditioning on the reference image, the same source instruction embedding $\bm{x}_{T_r}$ is associated with a multi-modal conditional target distribution $p(\bm{x}_{T_t}\mid \bm{x}_{T_r})$.

\noindent \textbf{Proposition 1 (Mean Collapse).}  \textit{For an unconditioned flow, the marginal velocity field ${v}^\star(x_t)$ deterministically transports $x_{T_r}$ to the semantic expectation:}
\begin{equation}
    x_{T_r} + \int_0^1 {v}^\star(x_t) \mathrm{d}t = \mathbb{E}[x_{T_t}|x_{T_r}] 
    \label{eq:mean_collapse}
\end{equation}
Therefore, when $p(\bm{x}_{T_t}\mid \bm{x}_{T_r})$ is multi-modal, $\mathbb{E}[x_{T_t}|x_{T_r}] \notin \{x_{T_t}^1, \dots, x_{T_t}^K\}$, which induces a mean-collapse tendency and may prevent recovery of the correct target.
% the posterior $p(\bm{x}_{T_t}\mid \bm{x}_t,t)$ remains multi-modal and the FM-optimal velocity $\mathbb{E}\!\left[\bm{x}_{T_t}-\bm{x}_{T_r}\mid \bm{x}_t,t\right]$ averages multiple feasible target directions. This induces a mean-collapse tendency and may prevent recovery of the correct target.

\noindent \textbf{Semantic Transport via Conditioning.} 
To disambiguate the transport direction, we condition the transport on the reference image embedding $\bm{x}_{I_r}$.
We construct the conditional probability path to establish exact point-to-point semantic transport as:
\begin{equation}
    p_t(x_t|x_1 = x_{T_t}^i, x_{I_r}^i) = 
    \begin{cases} 
        \delta(x_t - x_{T_r}^i), & t = 0 \\ 
        \delta(t x_{T_t}^i + (1 - t)x_{T_r}^i), & t \in (0,1] 
    \end{cases}
    \label{eq:conditional_path}
\end{equation}
Subsequently, a corresponding velocity $\bm{v}^\star(\bm{x}_t|\bm{x}_{T_t}^i, \bm{x}_{I_r}^i)$ is defined to generate this path $p_t$, and we introduce the following proposition:
% \noindent \textbf{Assumption 1 (Conditional identifiability).}
% \textit{
% For any fixed reference condition $\bm{x}_{I_r}$ and $t\in(0,1]$, the target is identifiable from the intermediate state, i.e.,
% \begin{equation}
% p(\bm{x}_{T_t}^i\mid \bm{x}_t^i,\bm{x}_{I_r}^i)=\delta(t x_{T_t}^i + (1 - t)x_{T_r}^i).
% \end{equation}
% Equivalently, under the reference condition, each intermediate point $\bm{x}_t$ corresponds to a unique target embedding $\bm{x}_{T_t}$.
% }

\noindent \textbf{Proposition 2 (Semantic Transport).} \textit{Under above assumption, 
the marginal velocity field is mathematically identical to the exact conditional velocity field:}
\begin{equation}
    \bm{v}^{\star}(\bm{x}_t,t,\bm{x}_{I_r}) = \bm{v}^\star(\bm{x}_t,t|\bm{x}_{I_r}^i, \bm{x}_{T_t}^i)
    \label{eq:marginal_identity}
\end{equation}
Consequently, integrating this velocity along the path yields exact point-to-point semantic transport from $x_{T_r}^i$ strictly towards its corresponding target $x_{T_t}^i$:
\begin{equation}
\bm{x}_{T_t}
=
\bm{x}_{T_r}
+
\int_0^1 {\bm{v}}^{\star}(\bm{x}_t,t,\bm{x}_{I_r})\,\mathrm{d}t.
\label{eq:exact_semantic_transport}
\end{equation}
We provide the detailed proof in the supplementary material.
This suggests that, under the idealized conditional path assumption, reference conditioning can reduce multimodal ambiguity and support point-to-point semantic alignment.

\section{Experiments}
\subsection{Setups and Implementations}
\textbf{Datasets.}
Following prior zero-shot CIR works~\cite{agnolucci2024isearle,baldrati2023zero,saito2023pic2word}, we evaluate on three standard benchmarks: CIRR~\cite{CIRR}, CIRCO~\cite{baldrati2023zero} and Fashion-IQ~\cite{fashionIQ}.
CIRR consists of real-world image pairs with relative captions describing the desired transformation from a reference to a target.
CIRCO further scales composed retrieval to diverse everyday objects and scenes with more challenging distractors.
Fashion-IQ is an e-commerce fashion benchmark, where queries pair a product image with a brief modification text to retrieve the edited item.

\noindent \textbf{Implementation Details.}
Following prior zero-shot CIR works, we use CLIP~\cite{clip} as the frozen vision--language backbone and report results with two variants: ViT-B/32 and ViT-L/14.
For the flow matching model, we adopt the lightweight network design from MAR~\cite{li2024autoregressive}, implemented as a deep residual MLP with timestep conditioning.
We train our model on the HQ-Edit-200k~\cite{hui2024hq} synthetic image-editing dataset, which contains reference images paired with editing instructions, the corresponding edited results, and associated textual descriptions, matching the supervision signals required by our transport-based formulation.
For multi-negative steering, we first apply a rule-based strategy to identify negation-related instructions by checking whether the query contains keywords such as \texttt{no}, \texttt{without}, \texttt{remove}, and \texttt{replace}. For the detected cases, we further use a lightweight language model, TinyLlama-1.1B-Chat-v1.0~\cite{zhang2024tinyllama}, to decouple the negative component from the original instruction, which is then used to construct the additional negative steering signal at inference time.
Unless otherwise specified, all experiments are conducted on a single NVIDIA RTX 3090 GPU with $24$GB memory.
Further details on datasets, optimization, and hyperparameters are deferred to the supplementary material.

\subsection{Main Results}

\begin{table}[!t]
\centering
\caption{Comparison on CIRR (R@1/5/10) and CIRCO (mAP@5/10/25/50) under CLIP ViT-B/32 and ViT-L/14 backbones, including training cost. The best result of each category is shown in \textbf{bold}, and the second best is \underline{underlined}.}
\renewcommand{\arraystretch}{1.12}
\setlength{\tabcolsep}{4pt}
\resizebox{\textwidth}{!}{
\begin{tabular}{c l c c c c c c c c}
\toprule
\multirow{2}{*}{Backbone} & \multirow{2}{*}{Method} & \multirow{2}{*}{Training Costs}
& \multicolumn{3}{c}{CIRR}
& \multicolumn{4}{c}{CIRCO} \\
\cmidrule(lr){4-6}\cmidrule(lr){7-10}
& & & R@1 & R@5 & R@10 & mAP@5 & mAP@10 & mAP@25 & mAP@50 \\
\midrule

\multirow{8}{*}{\makecell{\rotatebox{90}{ViT-B/32}}}
& Image-only & -- & 6.7 & 23.0 & 59.2 & 1.5 & 1.9 & 2.3 & 2.6 \\
& Text-only  & -- & 21.8 & 45.2 & 57.4 & 2.5 & 2.6 & 2.9 & 3.1 \\
& PALAVRA~\cite{PALAVRA} & -- & 16.6 & 43.5 & 58.5 & 4.6 & 5.3 & 6.3 & 6.8 \\
& SEARLE~\cite{baldrati2023zero} & 8 A100, 4h & 24.3 & 53.3 & 66.1 & 8.9 & 9.4 & 10.6 & 11.2 \\
% & Slerp~\cite{jang2024spherical} & -- & 24.2 & 50.9 & 63.5 & 6.4 & 7.1 & 8.1 & 8.8 \\
& iSEARLE~\cite{agnolucci2024isearle} & A100, 12h & 25.2 & \underline{55.7} & \underline{68.1} & \underline{10.6} & \underline{11.2} & \underline{12.5} & \underline{13.3} \\
& SEARLE + CIG~\cite{wang2025generative} & NA & \underline{25.3} & 54.8 & \underline{68.1} & 10.2 & 10.6 & 11.8 & 12.5 \\
&\textcolor{gray}{MagicLens}~\cite{zhang2024magiclens}&\textcolor{gray}{64 TPU, 6h} & \textcolor{gray}{27.0}&\textcolor{gray}{58.0}&\textcolor{gray}{70.9}&\textcolor{gray}{23.1}&\textcolor{gray}{23.8}&\textcolor{gray}{25.8}&\textcolor{gray}{26.7}\\
& \cellcolor{my_green}Ours & \cellcolor{my_green}RTX 3090, 0.5h & \cellcolor{my_green}\textbf{25.5} & \cellcolor{my_green}\textbf{56.5} & \cellcolor{my_green}\textbf{69.8} & \cellcolor{my_green}\textbf{13.1} & \cellcolor{my_green}\textbf{13.4} & \cellcolor{my_green}\textbf{14.6} & \cellcolor{my_green}\textbf{15.3} \\
\midrule

\multirow{12}{*}{\makecell{\rotatebox{90}{ViT-L/14}}}
& Image-only & -- & 7.3 & 23.0 & 33.3 & 2.5 & 3.1 & 3.9 & 4.4 \\
& Text-only  & -- & 20.9 & 44.0 & 55.4 & 3.3 & 3.7 & 4.1 & 4.4 \\
& Pic2Word~\cite{saito2023pic2word} & 8 A100, 16h & 23.9 & 51.7 & 65.3 & 8.7 & 9.5 & 10.6 & 11.3 \\
& SEARLE~\cite{baldrati2023zero} & 8 A100, 4h & 24.2 & 52.4 & 66.3 & 11.7 & 12.7 & 14.3 & 15.1 \\
& Context-I2W~\cite{tang2024context} & 8 A100 24h & \underline{25.6} & \underline{55.1} & 68.5 & -- & -- & -- & -- \\
% & Slerp~\cite{jang2024spherical} & -- & 24.4 & 49.9 & 62.3 & 8.8 & 9.8 & 11.3 & 12.0 \\
& LinCIR~\cite{gu2024language} & 8 A100, 0.5h & 25.0 & 53.3 & 66.7 & 12.6 & \underline{13.6} & 15.0 & 15.9 \\
& LinCIR + CIG~\cite{wang2025generative} & NA & \underline{25.6} & 54.8 & 67.6 & \underline{13.0} & \underline{13.6} & 15.1 & 16.0 \\
& Compo-Diff~\cite{gu2023compodiff} & 128 A100, 231h & 18.2 & 53.1 & \textbf{70.8} & 12.6 & 13.4 & \underline{15.8} & \underline{16.4} \\
& iSEARLE~\cite{agnolucci2024isearle} & 12 & 25.4 & 54.1 & 67.5 & 11.3 & 12.7 & 14.5 & 15.3 \\
&\textcolor{gray}{MagicLens}~\cite{zhang2024magiclens}&\textcolor{gray}{128 TPU, 6h} & \textcolor{gray}{30.1}&\textcolor{gray}{61.7}&\textcolor{gray}{74.4}&\textcolor{gray}{29.6}&\textcolor{gray}{30.8}&\textcolor{gray}{33.4}&\textcolor{gray}{34.4}\\
&\textcolor{gray}{MCL (LLaMA2-7B)}~\cite{li2024improving}&\textcolor{gray}{NA} & \textcolor{gray}{26.2}&\textcolor{gray}{56.8}&\textcolor{gray}{70.0}&\textcolor{gray}{17.7}&\textcolor{gray}{18.9}&\textcolor{gray}{20.8}&\textcolor{gray}{21.7}\\ 
& \cellcolor{my_green}Ours & \cellcolor{my_green}RTX 3090, 0.5h & \cellcolor{my_green}\textbf{26.2} & \cellcolor{my_green}\textbf{56.1} & \cellcolor{my_green}\underline{68.6} & \cellcolor{my_green}\textbf{14.9} & \cellcolor{my_green}\textbf{15.7} & \cellcolor{my_green}\textbf{17.3} & \cellcolor{my_green}\textbf{18.2} \\
\bottomrule
\end{tabular}
}
\label{tab:cirr+circo}
\end{table}

\begin{table}[t]
\centering
\caption{Comparison on Fashion-IQ under CLIP ViT-B/32 and ViT-L/14 backbones, reported in Recall@10 and Recall@50 for Dress/Shirt/Toptee and their average.}
\renewcommand{\arraystretch}{1.15}
\setlength{\tabcolsep}{4pt}
\resizebox{0.7\textwidth}{!}{
\begin{tabular}{c l *{8}{c}}
\toprule
\multirow{2}{*}{Backbone} & \multirow{2}{*}{Methods}
& \multicolumn{8}{c}{Fashion-IQ} \\
\cmidrule(lr){3-10}
& &
\multicolumn{2}{c}{Dress} &
\multicolumn{2}{c}{Shirt} &
\multicolumn{2}{c}{Toptee} &
\multicolumn{2}{c}{Average} \\
\midrule

% =========================
% CLIP ViT-B/32
% =========================
\multirow{5}{*}{\makecell{CLIP-\\ViT-B/32}}
& Image-only & 3.9 & 10.8 & 7.5 & 14.0 & 6.2 & 13.4 & 5.9 & 12.7 \\
& Text-only & 13.6 & 31.8 & 20.3 & 35.3 & 20.2 & 40.5 & 18.0 & 35.9 \\
& PALAVRA~\cite{PALAVRA} & 17.3 & 35.9 & \underline{21.5} & 37.1 & 20.6 & 38.8 & 19.8 & 37.3 \\
& SEARLE~\cite{baldrati2023zero} & \underline{18.2} & \underline{38.6} & \textbf{24.8} & \textbf{41.1} & \underline{25.6} & \underline{46.2} & \underline{22.9} & \underline{42.0} \\
&\textcolor{gray}{MagicLens}~\cite{zhang2024magiclens}&\textcolor{gray}{21.5}&\textcolor{gray}{41.3}&\textcolor{gray}{27.3}&\textcolor{gray}{48.8}&\textcolor{gray}{30.2}&\textcolor{gray}{52.3}&\textcolor{gray}{26.3}&\textcolor{gray}{47.4}\\
& \cellcolor{my_green}Ours & \cellcolor{my_green}\textbf{24.8} & \cellcolor{my_green}\textbf{41.8} & \cellcolor{my_green}19.1 & \cellcolor{my_green}\underline{40.1} & \cellcolor{my_green}\textbf{26.4} & \cellcolor{my_green}\textbf{47.8} & \cellcolor{my_green}\textbf{23.4} & \cellcolor{my_green}\textbf{43.2} \\
\midrule

% =========================
% CLIP ViT-L/14
% =========================
\multirow{7}{*}{\makecell{CLIP-\\ViT-L/14}}
& Text-only & 18.3 & 30.1 & 13.6 & 30.0 & 17.4 & 33.9 & 16.4 & 31.3 \\
& Image-only & 10.7 & 19.9 & 4.5 & 12.2 & 8.4 & 16.5 & 7.8 & 16.2 \\
& Pic2Word~\cite{saito2023pic2word} & 20.0 & 40.2 & 26.2 & 43.6 & 27.9 & 47.4 & 24.7 & 43.7 \\
& SEARLE~\cite{baldrati2023zero} & 20.5 & 43.1 & 26.9 & 45.6 & 29.3 & 50.0 & 25.6 & 46.2 \\
& Context-I2W~\cite{tang2024context} & \underline{23.1} & \underline{45.3} & \textbf{29.7} & \textbf{48.6} & \underline{30.6} & \underline{52.9} & 27.8 & \textbf{48.9} \\
& LinCIR~\cite{gu2024language} & 20.9 & 42.4 & \underline{29.1} & \underline{46.8} & 28.8 & 50.2 & \underline{26.3} & \underline{46.5} \\
&\textcolor{gray}{MagicLens}~\cite{zhang2024magiclens}&\textcolor{gray}{25.5}&\textcolor{gray}{46.1}&\textcolor{gray}{32.7}&\textcolor{gray}{53.8}&\textcolor{gray}{34.0}&\textcolor{gray}{57.7}&\textcolor{gray}{30.7}&\textcolor{gray}{52.5}\\
& \cellcolor{my_green}Ours & \cellcolor{my_green}\textbf{31.6} & \cellcolor{my_green}\textbf{48.5} & \cellcolor{my_green}24.4 & \cellcolor{my_green}44.3 & \cellcolor{my_green}\textbf{33.2} & \cellcolor{my_green}\textbf{53.9} & \cellcolor{my_green}\textbf{29.7} & \cellcolor{my_green}\textbf{48.9} \\
\bottomrule
\end{tabular}
}
\label{tab:fashioniq}
\end{table}
Tab.~\ref{tab:cirr+circo} and Tab.~\ref{tab:fashioniq} compare \textsc{FlowCIR} with both \emph{textual-inversion-based} methods (including Pic2Word~\cite{saito2023pic2word}, SEARLE~\cite{baldrati2023zero}, LinCIR~\cite{gu2024language} and iSEARLE~\cite{agnolucci2024isearle}) and \emph{generative-based} methods (including Compo-Diff~\cite{gu2023compodiff} and CIG~\cite{wang2025generative}) on CIRR~\cite{CIRR}, CIRCO~\cite{baldrati2023zero}, and Fashion-IQ~\cite{fashionIQ} under CLIP ViT-B/32 and ViT-L/14 backbones, together with training cost.
Overall, our method achieves the strongest or highly competitive performance across benchmarks while requiring substantially less training than prior methods.
For completeness, we also include comparisons with methods relying on substantially stronger external resources, including MagicLens~\cite{zhang2024magiclens}, which is trained on 36.7M private triplets with several TPUs, and MCL~\cite{li2024improving}, which utilizes an external LLM.

\noindent \textbf{CIRR \& CIRCO.}
On the more challenging CIRR and CIRCO benchmarks, \textsc{FlowCIR} achieves the strongest or highly competitive performance among prior public-data textual-inversion and generative ZS-CIR baselines, while remaining competitive with methods that rely on substantially stronger external resources.
With CLIP ViT-B/32, our method improves over the strongest competing baseline by \textbf{1.4\%} on CIRR R@5 ($55.7\rightarrow 56.5$), and \textbf{23.6\%} ($10.6\rightarrow 13.1$) on CIRCO mAP@5.
The gains become more evident under CLIP ViT-L/14: compared with the previous best results, \textsc{FlowCIR} improves CIRR R@1 by \textbf{2.3\%} ($25.6\rightarrow 26.2$) and CIRCO mAP@5 by \textbf{14.6\%} ($13.0\rightarrow 14.9$).
Notably, these improvements are obtained with only 0.5 hours of training on a single GPU, compared with the reported multi-GPU training costs of prior inversion-based methods, and dramatically higher costs for generative approaches, which are typically reported on substantially larger $80$GB GPUs.
This confirms that our flow-based conditional semantic transport is not only more accurate than existing textual-inversion and generative baselines, but also markedly more training-efficient.

\noindent \break \textbf{Fashion-IQ.}
Tab.~\ref{tab:fashioniq} further demonstrates that \textsc{FlowCIR} generalizes to the e-commerce setting of Fashion-IQ~\cite{fashionIQ}.
Under CLIP ViT-B/32, \textsc{FlowCIR} improves the best prior method by \textbf{+2.2\%} ($22.9\rightarrow 23.4$) in average R@10 and \textbf{+2.9\%} ($42.0\rightarrow 43.2$) in average R@50.
Under CLIP ViT-L/14, it yields a \textbf{+6.8\%} ($26.3\rightarrow 29.7$) gain in average R@10 and matches the best prior R@50.
Although the relative gains are smaller than those on CIRR/CIRCO, we attribute this to the domain-specific and fine-grained nature of Fashion-IQ, where many queries focus on subtle local fashion attributes and thus leave less room for semantic transport to yield larger improvements.
Overall, \textsc{FlowCIR} remains consistently competitive and attains the best average performance.

\subsection{Diagnostic Analysis}

\begin{table*}[t]
\small
\centering
\caption{Ablation studies of different components in \textsc{FlowCIR} on CIRR~\cite{CIRR} and CIRCO~\cite{baldrati2023zero}. We evaluate the contributions of Conditional Flow Matching (CFM) and Multi-Negative Steering (Neg-Steering).}
\setlength\tabcolsep{5.0pt}
\resizebox{0.9\linewidth}{!}{
\begin{tabular}{ccc|ccc|cccc}
\toprule \hline
&CFM & Neg-Steering 
& \multicolumn{3}{c|}{CIRR} 
& \multicolumn{4}{c}{CIRCO} \\
&\S~\ref{subsec:cfm}&\S~\ref{subsec:multineg} & R@1 & R@5 & R@10 & mAP@5 & mAP@10 & mAP@25 & mAP@50 \\
\midrule
Text-Only&\xmark & \xmark & 20.9 & 44.0 & 55.4 & 3.3 & 3.7 & 4.1 & 4.4 \\
(1)&\xmark & \cmark & 21.1 & 45.7 & 58.0 & 2.5 & 2.8 & 3.2 & 3.4 \\
(2)&\cmark & \xmark & 26.1 & 53.8 & 67.0 & 13.5 & 14.2 & 15.7 & 16.6 \\
\cellcolor{my_green}\textsc{\textbf{FlowCIR}}&\cellcolor{my_green}\cmark & \cellcolor{my_green}\cmark & \cellcolor{my_green}\textbf{26.2} & \cellcolor{my_green}\textbf{56.1} & \cellcolor{my_green}\textbf{68.6} &\cellcolor{my_green} \textbf{14.9} & \cellcolor{my_green}\textbf{15.7} & \cellcolor{my_green}\textbf{17.3} & \cellcolor{my_green}\textbf{18.2} \\
\bottomrule
\end{tabular}}
\label{tab:ablation}
\end{table*}
\noindent \textbf{Component Analysis.}
We conduct ablations to quantify the contributions of the two key components in \textsc{FlowCIR}: Conditional Flow Matching (CFM) and inference-time Multi-Negative Steering (Neg-Steering).
As shown in Tab.~\ref{tab:ablation}, we report results on CIRR and CIRCO by progressively adding these components on top of a text-only baseline.
\textbf{First}, the text-only baseline provides only limited performance, and adding the multi-negative steering alone on top of the textual instruction (Row (1)) does not constitute a reliable solution.
Although it brings a slight improvement on CIRR, it degrades performance on CIRCO, suggesting that refining the text-side instruction alone is insufficient for accurate cross-modal retrieval.
\textbf{Second}, introducing \emph{Conditional Flow Matching} (Row (2)) yields a substantial improvement.
This clearly demonstrates the advantage of modeling composition as \emph{semantic transport}: rather than directly manipulating text features, the model learns a reference-conditioned flow that transports the instruction representation toward target semantics, resulting in much stronger cross-modal composition.
The particularly large gains on CIRCO, together with the consistent improvements on CIRR, confirm that flow matching is an effective mechanism for composing visual and textual signals in zero-shot retrieval.
\textbf{Third}, adding Multi-Negative Steering on top of CFM leads to further and consistent gains, yielding the full \textsc{FlowCIR} model with the best overall performance.
These results show that CFM is the main driver of performance, highlighting the effectiveness of flow-based semantic transport for cross-modal composition, while Multi-Negative Steering is complementary and further improves robustness by reducing ambiguity caused by negated semantics.

\begin{table*}[t]
\small
\centering
\caption{Hyperparameter analysis of \textsc{FlowCIR} on CIRR and CIRCO, including $\lambda$, the hard-negative ratio $K$, and the comparison between FM and direct regression.}
\setlength\tabcolsep{5.0pt}
\resizebox{0.9\linewidth}{!}{
\begin{tabular}{c|cc|cc|cc|cc}
\toprule \hline
\multirow{2}{*}{Param.} 
& \multicolumn{4}{c|}{CIRR} 
& \multicolumn{4}{c}{CIRCO} \\
& \multicolumn{2}{c}{Val} 
& \multicolumn{2}{c|}{Test}
& \multicolumn{2}{c}{Val} 
& \multicolumn{2}{c}{Test} \\
\cline{2-9}
& R@1 & R@10 & R@1 & R@10 & mAP@5 & mAP@25 & mAP@5 & mAP@25 \\
\midrule

$\lambda = 0.1$  & 25.4 & 69.3 & 25.9 & 67.2 & 10.2 & 13.1 & 10.6 & 12.5 \\
$\lambda = 0.3$  & \underline{26.2} & \textbf{69.7} & \textbf{26.4} & \underline{68.1} & 13.9 & 15.7 & 14.2 & 16.8 \\
\cellcolor{my_green}$\lambda = 0.5$  & \cellcolor{my_green}\textbf{26.6} & \cellcolor{my_green}\underline{69.6} & \cellcolor{my_green}\underline{26.2} & \cellcolor{my_green}\textbf{68.6} & \cellcolor{my_green}\underline{14.8} & \cellcolor{my_green}\underline{16.9} & \cellcolor{my_green}\underline{14.9} & \cellcolor{my_green}\underline{17.3} \\
$\lambda = 0.7$  & 26.1 & 69.5 & \underline{26.2} & \underline{68.5} & \textbf{14.8} & \textbf{17.0} & \textbf{15.2} & \textbf{17.6} \\
$\lambda = 0.9$  & 26.0 & 69.5 & 25.8 & 68.4 & 14.5 & 16.8 & 14.8 & 17.2 \\
\hline

$K = \frac{1}{16}B$ & \underline{26.1} & \underline{69.2} & \underline{25.8} & 65.8 & \underline{13.9} & \underline{15.7} & \underline{14.2} & \underline{16.7} \\
\cellcolor{my_green}$K = \frac{1}{8}B$ & \cellcolor{my_green}\textbf{26.6} & \cellcolor{my_green}\textbf{69.6} & \cellcolor{my_green}\textbf{26.2} & \cellcolor{my_green}\textbf{68.6} & \cellcolor{my_green}\textbf{14.8} & \cellcolor{my_green}\textbf{16.9} & \cellcolor{my_green}\textbf{14.9} & \cellcolor{my_green}\textbf{17.3} \\
$K = \frac{1}{4}B$ & 25.8 & 58.8 & 25.5 & \underline{67.6} & 13.5 & 15.2 & 13.7 & 16.0 \\
\hline

\textit{Regressor} & \underline{23.7} & \underline{67.6} & \underline{23.8} & \underline{66.1} & \underline{11.9} & \underline{14.1} & \underline{12.3} & \underline{14.3} \\
\cellcolor{my_green}\textit{FM} & \cellcolor{my_green}\textbf{26.6} & \cellcolor{my_green}\textbf{69.6} & \cellcolor{my_green}\textbf{26.2} & \cellcolor{my_green}\textbf{68.6} & \cellcolor{my_green}\textbf{14.8} & \cellcolor{my_green}\textbf{16.9} & \cellcolor{my_green}\textbf{14.9} & \cellcolor{my_green}\textbf{17.3} \\
\bottomrule
\end{tabular}}
\label{tab:hyperparam}
\end{table*}
\noindent \textbf{Hyperparameter analysis.}
Table~\ref{tab:hyperparam} studies two important hyperparameters in \textsc{FlowCIR}, tuned on a held-out validation split: the weight $\lambda$ of the retrieval objective and the hard-negative ratio $K$, where $K$ is defined as the proportion of mined negatives relative to the minibatch size ($B$).
Overall, the method is stable across a broad range of settings, but moderate values consistently work best.
For $\lambda$, using a very small value (\eg, $\lambda=0.1$) leads to weaker CIRCO performance, indicating insufficient retrieval supervision, while overly large values do not bring further gains.
The best overall trade-off is achieved around $\lambda=0.5$, which yields the strongest or near-strongest results on both CIRR and CIRCO.
For hard-negative mining, selecting too few negatives (\eg, $K=\frac{1}{16}B$) or too many negatives (\eg, $K=\frac{1}{4}B$) is suboptimal, whereas a moderate choice $K=\frac{1}{8}B$ gives the best overall performance.
This suggests that an appropriate number of hard negatives is important: too few fail to provide sufficient discriminative pressure, while too many may introduce noisy distractors.

\noindent \textbf{Flows \textit{vs.} Regressor.}
We further analyze the rationale of our flow matching design by comparing it with a \emph{direct regressor} variant that uses the same model architecture and the same reference-image conditioning, but directly regresses the target residual instead of learning a time-dependent transport field.
As shown in the bottom part of Table~\ref{tab:hyperparam}, replacing flow matching with direct regression leads to a clear performance drop across all metrics, especially on CIRCO, where mAP@5 decreases from $14.8$ to $12.3$ on the test set.
In contrast, the flow-matching model consistently achieves the best results on both validation and test splits.
This comparison shows that learning a time-conditioned transport field is more effective and has stronger generalization ability than directly regressing the final embedding, likely because flow matching provides a more structured optimization target and better captures the semantic transition from instruction to target under the reference condition.

\noindent \textbf{Transport Target Space Analysis.}
We additionally study a key design choice in \textsc{FlowCIR}: whether the transport process should target the \emph{image} embedding space or the \emph{text} embedding space.
As shown in Tab.~\ref{tab:target_space_analysis}, transporting toward the target text embedding consistently outperforms transporting toward the target image embedding on both CIRR and CIRCO.
This comparison is closely related to prior generative-based methods such as Compo-Diff~\cite{gu2023compodiff}, which aim to conditionally generate target \emph{image} embeddings.
In contrast, our goal is \emph{semantic transport}: given the reference image and relative instruction, we transport the query toward the target \emph{text} embedding and then use this transported representation for retrieval.
We favor the text space for two reasons.
First, composed retrieval is inherently one-to-many, so supervising the model with a specific target image embedding may over-emphasize instance-level visual details, whereas target text embeddings provide a more semantics-centered signal.
Second, transporting from relative text to target text forms a more homogeneous mapping, which is easier to learn than a cross-space text-to-image transport.
These results support our design choice of modeling zero-shot CIR as text-side semantic transport rather than direct prediction in the image space.
\begin{table*}[t]
\small
\centering
\caption{Transport target space analysis of \textsc{FlowCIR} on CIRR and CIRCO.}
\setlength\tabcolsep{5.0pt}
\resizebox{0.9\linewidth}{!}{
\begin{tabular}{c|ccc|cccc}
\toprule \hline
Target Space
& \multicolumn{3}{c|}{CIRR}
& \multicolumn{4}{c}{CIRCO} \\
& R@1 & R@5 & R@10 & mAP@5 & mAP@10 & mAP@25 & mAP@50 \\
\midrule
Image Embedding & 23.8 & 50.4 & 62.2 & 12.7 & 13.0 & 14.3 & 15.0 \\
\cellcolor{my_green}Text Embedding
& \cellcolor{my_green}\textbf{26.2}
& \cellcolor{my_green}\textbf{56.1}
& \cellcolor{my_green}\textbf{68.6}
& \cellcolor{my_green}\textbf{14.9}
& \cellcolor{my_green}\textbf{15.7}
& \cellcolor{my_green}\textbf{17.3}
& \cellcolor{my_green}\textbf{18.2} \\
\bottomrule
\end{tabular}}
\label{tab:target_space_analysis}
\vspace{-10pt}
\end{table*}

\subsection{Qualitative Analysis}
\begin{figure}[t]
  \centering
  \includegraphics[width=1\linewidth]{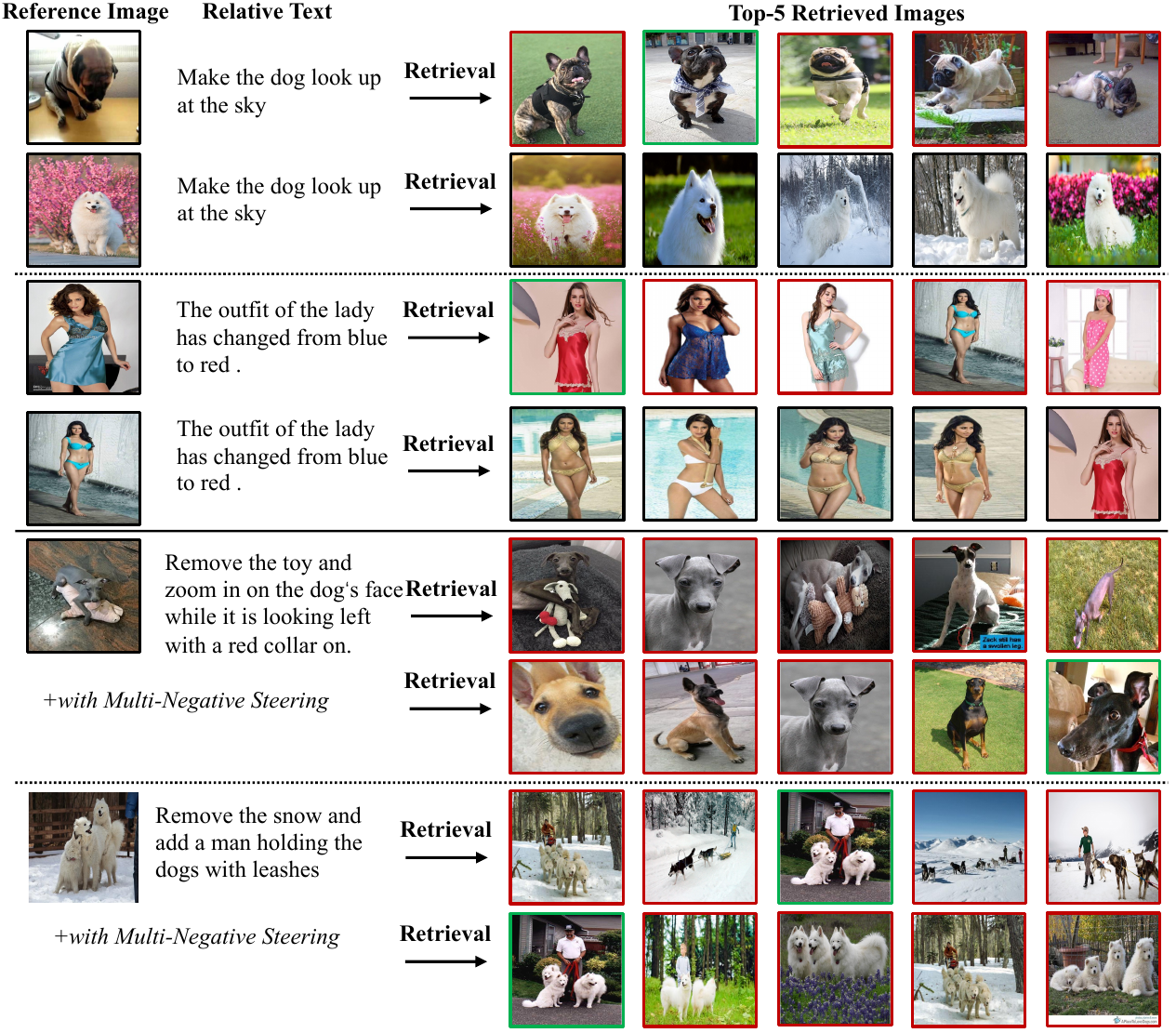}
  \caption{Qualitative retrieval results illustrating the effect of reference-conditioned transport and Multi-Negative Steering. Predicted targets are marked with {\color{green}\fbox{}} for correct retrievals and {\color{red}\fbox{}} for incorrect ones.}
  \label{fig:vis}
  \vspace{-10pt}
\end{figure}
\noindent \textbf{Effect of reference-conditioned transport.}
Fig.~\ref{fig:vis} illustrates that our method effectively composes cross-modal signals while preserving reference-image semantics, even under the same relative instruction.
In the first two examples, the instruction ``make the dog look up at the sky'' yields different retrieval behaviors depending on the reference: when the reference is a bulldog-like dog, the retrieved results largely preserve its breed and appearance while changing the pose; when the reference is a fluffy white dog, the retrieved images remain visually consistent with that appearance and mainly modify the head orientation.
A similar phenomenon appears in the third and fourth examples with the instruction ``the outfit of the lady has changed from blue to red,'' where the retrieved results preserve the person's pose and cloth style while modifying the clothing attribute.
In the fourth example, although there may be no exact target in the gallery, the retrieved images still remain close in clothing style and visual appearance, further showing that our model effectively combines the reference visual cues with the textual modification.
Overall, these examples show that \textsc{FlowCIR} enables effective cross-modal composition by jointly preserving the intended textual modification and the reference-image semantics; even under the same instruction, different reference conditions lead to distinct retrieval results.

\noindent \textbf{Effect of multi-negative steering.}
The bottom examples in Fig.~\ref{fig:vis} highlight the benefit of Multi-Negative Steering on negation-heavy instructions.
Without steering, retrieval often remains biased toward the negated concepts, reflecting the well-known difficulty of CLIP-style embeddings in separating affirmative and negated semantics.
For instance, for ``remove the toy and zoom in on the dog's face while it is looking left with a red collar on'', the unsteered results still often contain distracting toys or irrelevant objects, whereas applying Multi-Negative Steering yields more dog-face-centered results and better respects the intended removal.
Likewise, for ``remove the snow and add a man holding the dogs with leashes,'' the unsteered retrieval remains dominated by snowy scenes, while the steered version more consistently retrieves images with a human, leashes, and non-snowy environments.
These examples show that Multi-Negative Steering effectively pushes the composed representation away from undesired semantic directions, alleviating negation collapse and improving retrieval faithfulness.

\section{Conclusion}
In this paper, we introduce \textsc{\textbf{FlowCIR}}, a new paradigm for zero-shot composed image retrieval that formulates composition as conditioned semantic transport via flow matching in the embedding space.
Instead of reducing cross-modal composition to static token operations in text space, our method learns a lightweight conditional velocity field that transports the instruction representation toward a target-aligned query embedding under the guidance of the reference image.
We further propose an inference-only Multi-Negative Steering strategy that steers a negation-containing relative instruction away from its negated semantics, mitigating the limited negation handling of VLMs and improving robustness on negation-heavy queries.
Extensive experiments on three common benchmarks demonstrate that \textsc{FlowCIR} achieves highly competitive performance among recent ZS-CIR methods, while offering a lightweight training protocol.
We hope this work can inspire future research on flow-based compositional retrieval and open up a new paradigm for cross-modal retrieval beyond static composition.

\noindent \break \textbf{Acknowledgments.} 
This work was supported by National Natural Science Foundation of China (NSFC) Young Scientists Fund Category C (62402408), National Natural Science Foundation of China (NSFC) Young Scientists Fund Category B (62522216), Hong Kong SAR Research Grants Council (RGC) Early Career Scheme (26208924), and Hong Kong SAR Research Grants Council (RGC) General Research Fund (16219025).

\bibliographystyle{splncs04}
\bibliography{main}

\end{document}